\documentclass[10pt,journal,compsoc]{IEEEtran}
\ifCLASSOPTIONcompsoc
  \usepackage[nocompress]{cite}
\else
  \usepackage{cite}
\fi

\usepackage{amsmath,amssymb,color,amsmath, graphicx, float, subcaption, mathrsfs,color}
\usepackage[ruled]{algorithm2e} 
\usepackage{bm}
\usepackage{epstopdf}
\usepackage[font={small}]{caption}
\usepackage[font={footnotesize}]{subcaption}
\usepackage{cite}
\usepackage{amsthm}
\usepackage{url}

\newtheorem{thm}{Theorem}
\newtheorem{defn}{Definition}

\begin{document}
\title{An information-Theoretic Approach to Semi-supervised Transfer Learning}

\author{Daniel Jakubovitz,
David Uliel,
Miguel Rodrigues,
Raja Giryes
\IEEEcompsocitemizethanks{
\IEEEcompsocthanksitem Daniel Jakubovitz, David Uliel and Raja Giryes are with the School of Electrical Engineering, Tel Aviv University, Tel Aviv 69978, Israel. (email: daniel.jakubovitz@gmail.com, daviduliel@gmail.com, raja@tauex.tau.ac.il)
\IEEEcompsocthanksitem Miguel R. D. Rodrigues is with the Department of Electronic and Electrical Engineering, University College London, London, United Kingdom. (email: m.rodrigues@ucl.ac.uk)}
}


\IEEEtitleabstractindextext{
\begin{abstract}
Transfer learning is a valuable tool in deep learning as it allows propagating information from one "source dataset" to another "target dataset", especially in the case of a small number of training examples in the latter. Yet, discrepancies between the underlying distributions of the source and target data are commonplace and are known to have a substantial impact on algorithm performance.
In this work we suggest novel information-theoretic approaches for the analysis of the performance of deep neural networks in the context of transfer learning.
We focus on the task of semi-supervised transfer learning, in which unlabeled samples from the target dataset are available during network training on the source dataset.
Our theory suggests that one may improve the transferability of a deep neural network by incorporating regularization terms on the target data based on information-theoretic quantities, namely the Mutual Information and the Lautum Information. We demonstrate the effectiveness of the proposed approaches in various semi-supervised transfer learning experiments.
\end{abstract}

}

\maketitle
\IEEEdisplaynontitleabstractindextext
\IEEEpeerreviewmaketitle

\IEEEraisesectionheading{
\section{Introduction}
\label{sec:introduction}}
\IEEEPARstart{M}{achine} learning algorithms have lately come to the forefront of technological advancements, providing state-of-the-art results in a variety of fields \cite{Goodfellow16Deep}.
However, alongside their outstanding performance, these methods suffer from sensitivity to data discrepancies - any inherent difference between the training data and the test data may result in a substantial decrease in performance.
Moreover, to obtain good performance, a large amount of labeled data is necessary for their training. Such a substantial amount of labeled data is often either very expensive or simply unobtainable.

\begin{figure*}
    \centering
    \begin{subfigure}[c]{0.8\textwidth}
    \centering
    \includegraphics[width=\columnwidth]{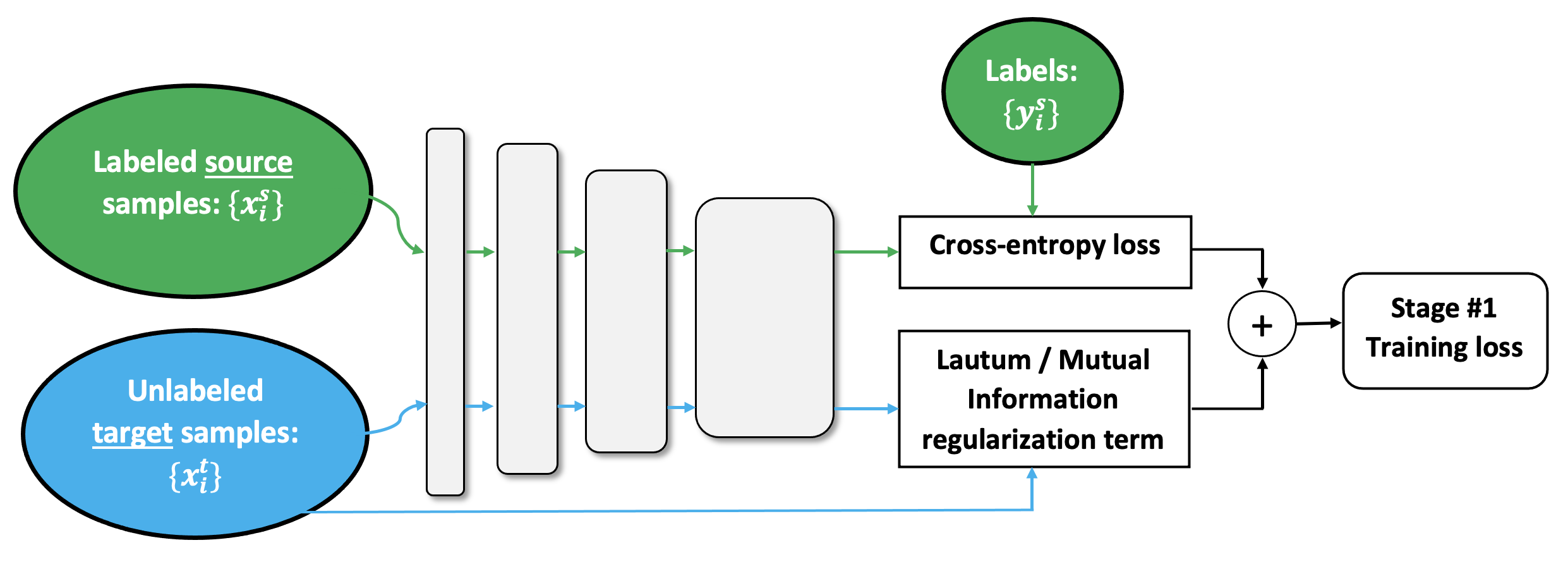}
    \caption{Pre-transfer training stage.}
    \end{subfigure}
    \hfill
    \begin{subfigure}[c]{0.8\textwidth}
    \centering
    \includegraphics[width=\columnwidth]{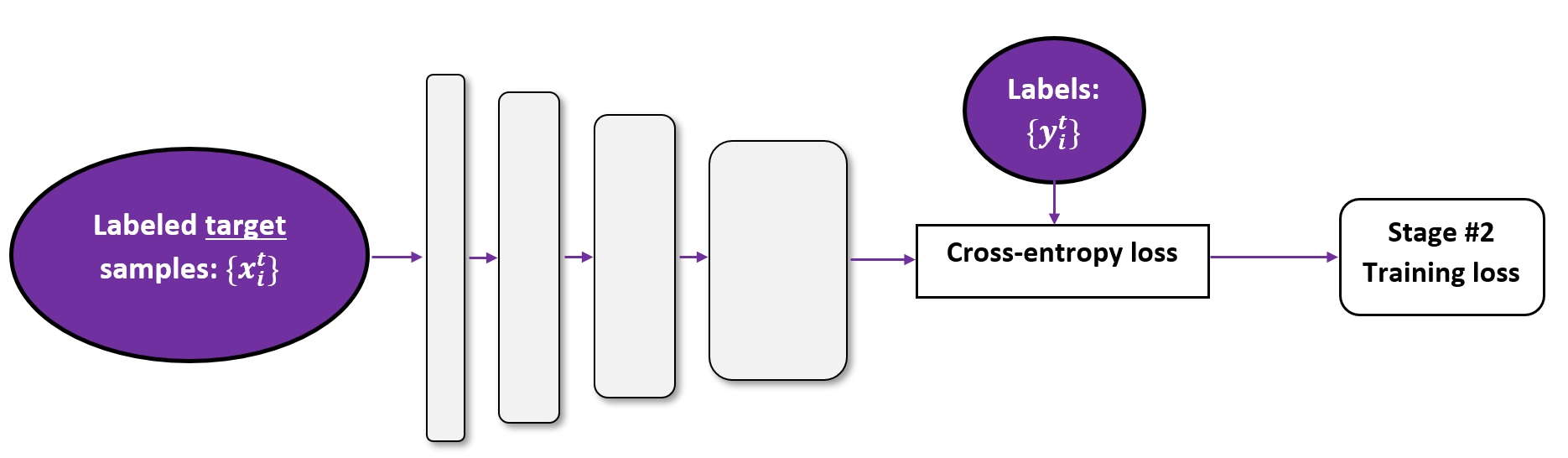}
    \caption{Post-transfer training stage.}
    \end{subfigure}
    \caption{Our semi-supervised transfer learning technique applies information-based regularization. Omitting the blue part in the Pre-transfer training stage (lower part of (a)), results in standard transfer learning.}
    \label{fig:Our_training_technique}
\end{figure*}

One popular approach to mitigate this issue is using "transfer learning", where training on a small labeled "target" dataset is improved by using information from another large labeled "source" dataset.
A common method for transfer learning uses the result of training on the source as initialization for training on the target, thereby improving performance on the latter \cite{Donahue_2014_Decaf}.

Transfer learning has been the focus of substantial research attention over the years. Plenty of different approaches have been proposed to encourage a more effective transfer from a source dataset to a target dataset, many of which aim at obtaining better system robustness to environment changes, so as to allow an algorithm to perform well even under some variations in the settings (e.g. changes in lighting conditions in computer vision tasks).
Several works take a more targeted approach and directly try to reduce the generalization error by reducing the difference in performance on specific source and target datasets \cite{Xuhong_2018_Explicit_TL}.
It is often the case that the target dataset has a large number of samples, though only a few of those samples are labeled. In this scenario a semi-supervised learning approach could prove to be beneficial by making effective use of the available unlabeled samples for training.


In this work we focus on the task of semi-supervised transfer learning.
We make a distinction between transfer learning and the related task of domain adaptation, where the former refers to the case of two sources of data which are different in both content and "styling" (e.g. the MNIST $\rightarrow$ SVHN case), whereas the latter refers to the case of two sources of data with the same content. Another relevant difference is that labeled data from the target distribution is typically available in the transfer learning case, yet no such labels are available in the domain adaptation case.

Plenty of works exist in the literature on transfer learning, semi-supervised learning and the use of information theory for the analysis of machine learning algorithms.
The closest work to ours is \cite{achille2018emergence} in which an information-theoretic approach is used in order to decompose the cross-entropy \emph{train} loss of a machine learning algorithm into several separate terms. However, unlike this work, we propose two different decompositions of the cross-entropy \emph{test} loss and make the relation to semi-supervised transfer learning.

{\bf Contribution.}
We consider the case of semi-supervised transfer learning in which plenty of labeled examples from a source distribution are available alongside just a few labeled examples from a target distribution; yet, we are also provided  with a large number of unlabeled samples from the latter.
This setup combines transfer learning and semi-supervised learning, where both aim at obtaining an improved performance on a target dataset with a small number of labeled examples.
In this work, we propose combining both methodologies to compound their advantages.
This setting represents the case where the learned information from a large labeled source dataset is used to obtain good performance when transferring to a mostly unlabeled target set, where the unlabeled examples of the target are available at training time on the source.
This scenario is commonplace in situations where obtaining labels for target data samples is an expensive or lengthy process, however labels for source data samples are of abundance and readily available, most often due to their open-source availability or an enhanced ease of their labeling.
An example scenario is when transferring from an open-source dataset (e.g. X-Ray images of a variety of human organs from a variety of machines) with plenty of available labeled samples, to a more specific target dataset (X-Ray images from a specific type of machine, a specific point of view, a specific human organ in view etc.) for which few labels are available.

We present two theoretical derivations that lead to novel semi-supervised transfer learning techniques.
We take an information-theoretic approach to examine the cross-entropy test loss of machine learning methods; we present two decompositions of the loss into several different terms that account for different aspects of its behavior.
These derivations lead to new regularization terms: (i) "Lautum regularization" which relies on the maximization of the Lautum information \cite{Verdu_08_Lautum} between unlabeled data samples drawn from the target distribution and the learned model weights, and (ii) "Mutual Information regularization", derived from the Information Bottleneck Theorem \cite{tishby2000information}, which relies on the minimization of the mutual information between the learned model parameters and the unlabeled target samples.

Figure~\ref{fig:Our_training_technique} provides a general illustration of our two approaches.
We corroborate their effectiveness with experiments of semi-supervised transfer learning for neural networks on image classification tasks.
We examined the transfer in two cases: from the MNIST dataset to the notMNIST dataset (which consists of the letters A-J in grayscale images) and from the CIFAR-10 dataset to 10 specific classes of the CIFAR-100 dataset.
We compare our results to three other methods: (1) Temporal Ensembling (TE) \cite{Laine_18_TemporalEnsembling}, a prominent method for semi-supervised training which we apply in a transfer learning setup. We examine TE both by itself and combined with Lautum regularization; (2) the Multi-kernel Maximum Mean Discrepancy (Mk-MMD) method \cite{Mk_MMD}, which is popular in semi-supervised transfer learning; (3) standard transfer learning which does not use any of the unlabeled samples.
We examine our two approaches both when separately applied (i.e. each approach is applied and analysed without the other), and when combined (i.e. when both approaches are applied at the same time within the same training context).
The advantages of our proposed methods are demonstrated in our experimental results which outperform the other compared methods.


\section{Related Work}
\label{sec:Related Work}
Plenty of works exist in the literature on transfer learning, semi-supervised learning and using information theory for the analysis of machine learning algorithms. We hereby overview the ones most relevant to our work. 

\textbf{Transfer learning.}
Transfer learning \cite{JialinPan_10_TransferSurvey, Tan_2018_DL_Transfer_Survey} is a useful training technique when the goal is to adapt a learning algorithm, which was trained on a source dataset, to perform well on a target dataset that is potentially very different in content compared to the source. This technique can provide a significant advantage when the number of training samples in the source dataset is large compared to a small number in the target, where the knowledge extracted from the source dataset may be relevant also to the target.

The work in \cite{Jason_2014_HowTransferable} relates to a core question in transfer learning: which layers in the network are general and which are more task specific, and precisely how transferability is affected by the distance between two tasks.
In a recent work \cite{Lampinen_19_TransferLinear} an analytical theory of how knowledge is transferred from one task to another in deep linear networks is presented. A metric is defined to quantify the amount of knowledge transferred between a pair of tasks.

Practical approaches to improving performance in transfer learning settings have been proposed in many works.
In \cite{Xuezhi_2014_Flexible_Transfer}, transfer learning in the context of regression problems is examined. A transfer learning algorithm that does not assume that the support of the target distribution is contained in the support of the source distribution is proposed. This notion leads to a more flexible transfer.
In \cite{Wei_18_Learning_to_transfer} the framework of "learning to transfer" is proposed in order to leverage previous transfer learning experiences for better transfer between a new pair of source and target datasets.
In \cite{Zamir_2018_Taskonomy} the structural relation between different visual tasks is examined in the feature space. The result is a taxonomic map that enables a more efficient transfer learning with a reduced amount of labeled data.
In \cite{Li2019_Delta} a regularized transfer learning framework that preserves the outer layer outputs of the target network (post-transfer) is proposed.
In \cite{NEURIPS2019_Negative_Transfer_Catastrophic_forgetting} the authors provide an empirical deep-dive into the inner workings of catastrophic forgetting and negative transfer, and propose the Batch Spectral Shrinkage approach to discourage the transfer of irrelevant spectral components of the learned representations.
In \cite{NEURIPS2020_autofocused_oracles} the authors propose the method of oracle Auto-focusing, which allows for a dynamic correction of data distribution shifts.

\textbf{Semi-supervised learning.}
Semi-supervised learning \cite{Zhu_06_SemiSupervisedSurvey} is typically used when there is little labeled data for training, yet more unlabeled data is available.
The literature on semi-supervised training is vast and describes a variety of techniques for performing effective semi-supervised learning that would make good use of the available unlabeled data in order to improve model performance.
Most of these techniques rely on projecting the relation between the available labeled samples and their labels onto the unlabeled samples and the model's predictions for them.

In \cite{Yves_05_SemiSupervised_Entropy} minimum entropy regularization is proposed. This technique modifies the cross-entropy loss used for training in order to encourage a deep neural network to make confident predictions on unlabeled data.
In \cite{HausserMC_17_LearningByAssociation}, a new framework for semi-supervised training of neural networks called "associative learning" is proposed. In this framework "associations" are made between the embeddings of the available labeled data and the unlabeled data. An optimization process is then used to encourage correct "associations", which make better use of the unlabeled data.
In \cite{Haitian_18_SemiSuperEntropyConstraints}, a method is proposed for combining several different semi-supervised learning techniques using Bayesian optimization.
In \cite{Papernot_2017_SemiSuperKnowledge}, a semi-supervised framework that allows labeled training data privacy is proposed. In this framework, knowledge is transferred from teacher models to a student model in a semi-supervised manner, thereby precluding the student from gaining access to the labeled training data which is available to the teachers.
In \cite{Oliver_2018_RealisticSSL}, various semi-supervised learning algorithms are evaluated on real-world applications, yet no specific attention is paid to the transfer learning case and the effects of fine-tuning a pre-trained network.

Two notable works that employ semi-supervised training techniques are \cite{Neal_18_SemiSupervised_Minimizing_Variance} and \cite{Abhishek_18_SemiSuper_GANs}.
In \cite{Neal_18_SemiSupervised_Minimizing_Variance}, a semi-supervised deep kernel learning model is presented for regression tasks. 
In \cite{Abhishek_18_SemiSuper_GANs}, a GAN-based method is presented. It is proposed to estimate the tangent space to the learned data manifold using GANs, infer the relevant invariances and then inject these into the learned classifier during training.

Other prominent works that focus on semi-supervised transfer learning are \cite{Gupta_2018_SemiSupervisSentiment} which examines semi-supervised transfer learning for sentiment classification, \cite{Zhou_2018_SSL_TL} where semi-supervised transfer learning is examined for different training strategies and model choices, and \cite{CVPR2021_SSTL} where an adaptive consistency regularization paradigm is suggested to optimize performance.

All of the above mentioned works differ from ours in their fundamental approach - they address the discrepancy between the source and target data either in the input space or in the feature space, yet disregard the effect of the chosen loss function and its impact on the mitigation of this discrepancy.
In contrast, our work is focused on the mathematical analysis of cross-entropy loss which is commonly used in classification tasks. This provides our approach with a competitive advantage compared to other prominent methods.


\textbf{Information theory and machine learning.}
Information theory has largely been used to provide theoretical insight into the intricacies of machine learning algorithms.
In \cite{Tishby99informationBotteleneck}, the Information Bottleneck framework has been presented. This framework formalizes the tradeoff between algorithm sufficiency (fidelity) and complexity. It has been analyzed in various works such as \cite{Shamir_10_InformationBottleneckAnalysis} and \cite{Friedman_01_Multivariate_IB}. Following works \cite{Tishby15_information_deeplearning, Shwartz-ZivT17_blackbox} explicitly apply information bottleneck principles to deep learning.
In \cite{Andrew_18_OnTheInformationBottleneck} several claims from \cite{Shwartz-ZivT17_blackbox} are examined and challenged.
In \cite{Marylou_18_MI_in_DL} useful methods for computing information-theoretic quantities are proposed for several deep neural network models.
In \cite{Gibbs_2022} the authors provide an information-theoretic analysis of the generalization ability of Gibbs-based transfer learning algorithms. They show how the benefits of transfer learning can be explained by the framework of the bias-variance tradeoff.

In order to utilize information-theoretic quantities such as the Mutual Information (MI) in neural network training, efficient estimation methods are required. Recent works harness the strength of DNNs in order to estimate the MI.
The MINE approach \cite{belghazi2018mine} suggests using a DNN for estimating a lower bound for the MI. The work in \cite{poole2019variational} compared several variational methods of MI estimation and presented their bias-variance tradeoff.
The work in \cite{song2019understanding} analyzed the limitations of MI estimators and suggested a way to modify MINE for reducing its variance.


The closest work to ours is \cite{achille2018emergence}, where an information-theoretic approach is used to decompose the cross-entropy loss of a machine learning algorithm into several separate terms. It is suggested that overfitting the training data is mathematically encapsulated in the mutual information between the training data labels and the learned model weights; i.e. the mutual information essentially represents the ability of a neural network to \emph{memorize} the training data \cite{GeneralizationError_18_Jakubovitz}.
Consequently, a regularizer that alleviates the phenomenon of overfitting is proposed, and initial results of its efficiency are presented.
However, we propose two different decompositions of the cross-entropy \emph{test} loss and make the connection to semi-supervised transfer learning.
We propose two regularizers: (i) based on the Lautum information between the model weights and the unlabeled samples of the target dataset; (ii) based on the mutual information between the model weights and the unlabeled samples of the target dataset.
Both of these lead to an improved semi-supervised transfer learning technique, and yield experimental results that corroborate our theoretical analysis.

\section{The cross-entropy loss - an information theory perspective}
\label{sec:The Cross-Entropy Loss - an Information Theory Perspective}

Let $\mathcal{D} =\{ (x_i, y_i) \}_{i=1}^{N}$ be a fixed training set with $N$ training samples that is used to train a learning algorithm with a set of weights $w$.
Given $\mathcal{D}$, the learning algorithm selects a specific hypothesis from the hypothesis class according to the distribution $p(w_\mathcal{D})$. In the case of neural networks, selecting a hypothesis is equivalent to training the network on the data.

We denote by $w_{\mathcal{D}}$ the model weights that were learned using the fixed training set $\mathcal{D}$, and by $f(y|x,w_{\mathcal{D}})$ the learned classification function which given the weights $w_{\mathcal{D}}$ and a $D$-dimensional input $x \in \mathbb{R}^D$ computes the probability of the $K$-dimensional label $y \in \mathbb{R}^K$.
Note that we treat $f(y|x,w_{\mathcal{D}})$ as a probability distribution over the random variables $y,x,w_{\mathcal{D}}$, where $\mathcal{D}$ is a fixed parameter.

The learned classification function is then tested on data drawn from the true underlying distribution $p(x,y)$.
Ideally, as a result of the training process the learned classification function $f(y|x,w_{\mathcal{D}})$ would highly resemble the ground-truth label distribution $p(y|x)$, and similarly $f(x,y|w_{\mathcal{D}})$ would highly resemble $p(x,y)$.
In practice, the learned function does not perfectly match the real distribution although this is the goal of training.

Using these notations, we analyze the cross-entropy loss between the predictions of a learned classification function and the true label distribution, which is commonly employed for training in classification tasks. In our derivations we used several information-theoretic measures which we present hereafter.
Let $X,Y$ be two random variables with respective probability density functions $p(x), p(y)$ and the joint probability density function $p(x,y)$.
The following are the definitions of three information-theoretic measures which are relevant to our derivations.

\begin{defn}[Mutual information]
\label{def:MI}
The mutual information (MI) between $X$ and $Y$ is defined by
\begin{equation}
    I(X;Y) = \iint p(x,y) \log \left\{ \frac{p(x,y)}{p(x)p(y)} \right\} dx dy.
\end{equation}
\end{defn}

The MI captures the dependence between two random variables. It is the Kullback-Leibler divergence between the joint distribution and the product of the marginal distributions.

\begin{defn}[Lautum information]
\label{def:Lautum}
The Lautum information between $X$ and $Y$ is
\begin{equation}
    L(X;Y) = \iint p(x)p(y) \log \left\{ \frac{p(x)p(y)}{p(x,y)} \right\} dx dy.
\end{equation}
\end{defn}

This measure is the Kullback-Leibler divergence between the product of the marginal distributions and the joint distribution.
Similar to the MI, the Lautum information is related to the dependence between two random variables. However, it has different properties than the MI, as outlined in \cite{Verdu_08_Lautum}.

\begin{defn}[Differential entropy]
\label{def:entropy}
The differential entropy of a random variable $X$ is defined by
\begin{equation}
    H(X) = - \int p(x) \log p(x) dx.
\end{equation}
\end{defn}

\noindent {\bf Main theoretical results.} Having these definitions, we present our main theoretical results in the two following theorems:

\begin{thm}
\label{thm:test_ce_loss}
For a classification task with a ground-truth distribution $p(y|x)$, a fixed training set $\mathcal{D}$, learned weights $w_\mathcal{D}$ and learned classification function $f(y|x, w_\mathcal{D})$, the expected cross-entropy loss of a machine learning algorithm on the test distribution is approximately
\begin{equation}
    \label{eq:thm_test_ce_loss}
    \mathbb{E}_{w_{\mathcal{D}}} \left\{ KL( p(x,y) || f(x,y|w_{\mathcal{D}}) ) \right\} + H(y|x) - L(w_{\mathcal{D}};x),
\end{equation}
where the expected cross-entropy test loss is defined by
\begin{equation}
    \label{eq:Test_CE_appendix}
    \mathbb{E}_{(x,y) \sim p(x,y)} \mathbb{E}_{w \sim p(w_\mathcal{D})} \{-\log f(y|x,w_\mathcal{D})\}.
\end{equation}
\end{thm}
Note that $KL$ signifies the Kullback-Leibler divergence and that we treat the training set $\mathcal{D}$ as a fixed parameter, whereas $w_{\mathcal{D}}$ and the examined test data $(x,y)$ are treated as random variables.
We refer the reader to Appendix~A for the proof of Theorem~\ref{thm:test_ce_loss}.

In accordance with Theorem~\ref{thm:test_ce_loss}, the three terms that compose the expected cross-entropy test loss represent three different aspects of the loss of a learning algorithm performing a classification task:

\begin{itemize}
\item {\bf Classifier mismatch}:
$\mathbb{E}_{w_{\mathcal{D}}} \left\{ KL \left( p(x,y) || f(x,y | w_{\mathcal{D}}) \right) \right\}$, measures the deviation of the learned classification function's data distribution $f(x,y|w_{\mathcal{D}})$ from the true distribution of the data $p(x,y)$.
It is measured by the KL-divergence, which is averaged over all possible instances of $w$ parameterized by the fixed training set $\mathcal{D}$. This term essentially measures the ability of the weights learned from $\mathcal{D}$ to represent the true distribution of the data.

\item {\bf Intrinsic Bayes error}:
$H(y|x)$, represents the inherent uncertainty of the labels given the data samples.

\item {\bf Lautum information}:
$L(w_\mathcal{D}; x)$, represents the dependence between $w_{\mathcal{D}}$ and $x$. Note that $L(w_\mathcal{D}; x) = \mathbb{E}_{w_{\mathcal{D}}} \{ KL( p(x) || p(x|w_{\mathcal{D}}) ) \}$, it essentially measures how much $p(x|w_{\mathcal{D}})$ deviates from $p(x)$ on average over the possible values of $w_{\mathcal{D}}$.
\end{itemize}

Our formulation suggests that a machine learning algorithm, which is trained relying on empirical risk minimization, implicitly aims to maximize the Lautum information $L(w_\mathcal{D};x)$ in order to minimize the cross-entropy loss.
At the same time, the algorithm aspires to minimize the KL-divergence between the ground-truth distribution of the data and the learned classification function.
The intrinsic Bayes error cannot be minimized and remains the inherent uncertainty in the task.
Namely, the formulation in \eqref{eq:thm_test_ce_loss} suggests that promoting a larger Lautum information between the data samples and the learned model weights would be beneficial for reducing the model's test error on unseen data drawn from $p(x,y)$.

We next present a theorem that relates the classification error to the MI rather than the Lautum information.

\begin{thm}
\label{thm:test_ce_loss_2}
For a classification task with a ground-truth distribution $p(y|x)$, fixed training set $\mathcal{D}$, learned weights $w_\mathcal{D}$ and learned classification function $f(y|x, w_\mathcal{D})$, the expected cross-entropy loss of a machine learning algorithm on the test distribution is equal to
\begin{align}
\begin{split}
    & \mathbb{E}_{x, w_{\mathcal{D}}} \left\{ KL( p(y|x,w_{\mathcal{D}}) || f(y|x,w_{\mathcal{D}}) ) \right\} + \\
    & H(x,y|w_{\mathcal{D}}) - H(y) + I(x;w_{\mathcal{D}}),
\end{split}
\end{align}
where the expected cross-entropy test loss is defined by
\begin{equation}
    \label{eq:Test_CE_appendix}
    \mathbb{E}_{(x,y) \sim p(x,y)} \mathbb{E}_{w \sim p(w_\mathcal{D})} \{-\log f(y|x,w_\mathcal{D})\}.
\end{equation}
\end{thm}

The interplay between the different terms in Theorem~\ref{thm:test_ce_loss_2} sheds light on several aspects of the task's expected cross-entropy test loss:
\begin{itemize}
    \item \textbf{Conditional entropy}: $H(x,y| w_{\mathcal{D}})$, a measurement of the model weights' capability to capture the data distribution.
    \item \textbf{Label entropy}: $H(y)$, the intrinsic uncertainty in the labels.
    \item \textbf{Mutual information (MI)}: $I(x; w_{\mathcal{D}})$, an overfitting term, which represents the memorization of the inputs by the model weights.
    \item \textbf{Model inefficiency}: represented by the KL divergence term and relates to the discrepancy between the learned model classification function $f(y|x,w_{\mathcal{D}})$ and the true underlying distribution $p(y|x,w_{\mathcal{D}})$.
\end{itemize}
We refer the reader to Appendix~B for the proof of Theorem~\ref{thm:test_ce_loss_2}. Note that our theorem, unlike in \cite{achille2018emergence}, uses the MI term to make the relation to the transfer learning task.

\section{Regularized information in Semi - Supervised transfer learning }
\label{sec:regularized_information_in_semi_supervised_transfer_learning}
We turn to show how our theory may be applied to the task of semi-supervised transfer learning.
In standard transfer learning, which consists of a pre-transfer stage and a post-transfer stage, a neural network is trained on a labeled source dataset and then fine-tuned on a smaller labeled target dataset.
In semi-supervised transfer learning, which we study here, we assume that an additional large set of unlabeled examples from the target distribution is available during training on the source data.

Semi-supervised transfer learning is highly beneficial in scenarios where the available target dataset is only partially annotated. Making use of the unlabeled part of this dataset, which is usually substantially bigger than the labeled part, holds the potential to considerably improve the obtained performance. Thus, if this unlabeled part is a-priori available, then using it from the beginning of training can potentially improve the obtained performance results.

In order to use the unlabeled samples of the target dataset during pre-transfer training on the source dataset, we leverage the formulations in Theorem~\ref{thm:test_ce_loss}, and Theorem~\ref{thm:test_ce_loss_2}.
Both decompositions contain a term which relates to the information that the neural network weights hold with respect to the input samples.
By introducing Lautum or MI regularizations in the pre-transfer training stage, we promote a performance improvement on the target dataset.

\subsection{Lautum Information regularization in semi-supervised transfer learning}
\label{sec:Lautum information based semi-supervised transfer learning}



When inspecting the decomposition we obtained in Theorem~\ref{thm:test_ce_loss}, it is clear that by using unlabeled samples the classifier mismatch term cannot be minimized due to the lack of labels;
the intrinsic Bayes error is a characteristic of the task and cannot be minimized either; yet, the Lautum information does not depend on the labels and can therefore be calculated and maximized.

When the Lautum information is calculated between the model weights and data samples drawn from the target distribution, its maximization would encourage the learned weights to better relate to these samples, and by extension to better relate to the underlying probability distribution from which they were drawn.
Therefore, it is expected that an enlarged Lautum information will yield an improved performance on the target test set.
We accordingly aim at maximizing $L(w_{\mathcal{D}};x)$ during training. The pre-transfer maximization of the term $L(w_{\mathcal{D}};x)$, which is computed with samples drawn from the target distribution, would make the learned weights more inclined towards good performance on the target set right from the beginning. At the same time, the cross-entropy loss at this stage is calculated using labeled samples from the source dataset.
In the post-transfer stage, the cross-entropy loss is calculated using labeled samples from the target dataset, and therefore $L(w_{\mathcal{D}};x)$ is implicitly maximized during this stage as well.
We have empirically observed that explicitly maximizing the Lautum information between the unlabeled target samples and the model weights during post-transfer training by imposing Lautum regularization in addition to (or instead of) during pre-transfer training does not lead to improved results.

To summarize, our semi-supervised transfer learning approach optimizes two goals at the same time:
(i) minimizing the classifier mismatch $\mathbb{E}_{w_{\mathcal{D}}} \left\{ KL \left( p(x, y) || f(x, y | w_{\mathcal{D}}) \right)\right\}$, which is achieved using the labeled data both for the source and the target datasets during pre-transfer and post-transfer training respectively; and
(ii) maximizing the Lautum information $L(w_{\mathcal{D}};x)$, which is achieved explicitly using the unlabeled target data during pre-transfer training by imposing Lautum regularization, and in the post-transfer stage implicitly through the minimization of the cross-entropy loss which is evaluated on the labeled target data.
Figure~\ref{fig:Our_training_technique} summarizes our training scheme.

\noindent {\bf Estimating the Lautum information.}
We are interested in using the Lautum information as a regularization term, which we henceforth refer to as "Lautum regularization".
Since computing the Lautum information between two random variables requires knowledge of their probability distribution functions (which are high-dimensional and hard to estimate), we assume that $w_{\mathcal{D}}$ and $x$ are jointly Gaussian.\footnote{Note that this assumption is not very plausible for the distribution of the input samples, however could be reasonably plausible for the weights of the learned model}
This assumption allows us to explicitly compute the terms that make up the Lautum information, in addition to providing relatively simple computations and good experimental results as shown in Section~\ref{sec:Experiments}.

Since we only have one instance of the network weights at any specific point during training, and due to the computational complexity of using the network weights to estimate the Lautum information, we make use of the network's output logits to obtain an approximation for the model weights.
Essentially, the estimation is of the Jacobian matrix, i.e. the gradients of the learned classifier output with respect to the input.
An intuition to this approximation may be obtained from the linear case of $f=Wx$, where the derivative of the output with the respect to the input is $W = \frac{\partial{f}}{\partial{x}}$.
Estimating the Jacobian matrix comes with a high computational cost as the matrix is large in any practical scenario. Therefore, in order to perform more effective training and reduce the required resources, we draw inspiration from the approaches described in \cite{hoffman2019robust, NEURIPS2021_Sliced_MI}, and use random projections of the classifier output logits to a scalar value, of which we then take the derivative with respect to the inputs in order to estimate the Jacobian. In this way we are able to efficiently use the network outputs (logits) as a proxy for its weights.


In practice, to randomly project the logits to a scalar, we draw a random vector from a normal distribution with zero mean and variance $\frac{1}{\sqrt{K}}$ where $K$ represents the dimension of the logits vector. Then, we calculate the inner product between the classifier output logits and the random projection vector. Empirically, we noticed that by drawing the random projection vector once every epoch and using it throughout the iterations in the epoch, we obtain a more stable training and eventually achieve better results.


As shown in \cite{Verdu_08_Lautum}, the Lautum information between two jointly Gaussian random variables $(w,x)$ with covariance 
\begin{align}
\left[
\begin{array}{cc}
     \Sigma_{w} & \Sigma_{wx} \\
     \Sigma_{xw} & \Sigma_{x}
\end{array}
\right],
\end{align}
where $\Sigma_{x} \succ 0$ and $\Sigma_{w} \succ 0$, is given by
\begin{align}
\label{eq:Lautum_Gaussian}
\begin{split}
L(w;x) = & \log \left\{\det(I - \Sigma_{x}^{-1} \Sigma_{xw} \Sigma_{w}^{-1} \Sigma_{wx}) \right\} \\
& + 2 tr ((I - \Sigma_{x}^{-1} \Sigma_{xw} \Sigma_{w}^{-1} \Sigma_{wx})^{-1} - I).
\end{split}
\end{align}
where $det$ represents the matrix Determinant, and $tr$ is the matrix Trace.

The covariance matrix of our target dataset $\Sigma_{x}$ is evaluated once before training using the entire target training set, whereas $\Sigma_{w}, \Sigma_{wx}, \Sigma_{xw}$ are evaluated during training using the current mini-batch in every iteration.
All these matrices are estimated using standard sample covariance estimation based on the current mini-batch in every iteration, e.g.
\begin{align}
\Sigma_{x} = \frac{1}{N_{batch}} \sum_{i=1}^{N_{batch}} (x_i - \mu_x) (x_i - \mu_x)^T,
\end{align}
and
\begin{align}
\Sigma_{xw} = \frac{1}{N_{batch}} \sum_{i=1}^{N_{batch}} (x_i - \mu_x) (w_i - \mu_w)^T,
\end{align}
where $$\mu_x = \frac{1}{N_{batch}} \sum_{i=1}^{N_{batch}} x_i, ~~~~~~ \mu_{w} = \frac{1}{N_{batch}} \sum_{i=1}^{N_{batch}} w_i$$ represent the sample mean values of $x$ and $w$ respectively (i.e. their average values in the current mini-batch) and $N_{batch}$ denotes the mini-batch size.
The dimensions of these matrices are $\Sigma_{x} \in \mathbb{R}^{D \times D}, \Sigma_{w} \in \mathbb{R}^{K \times K}, \Sigma_{xw} \in \mathbb{R}^{D \times K}, \Sigma_{wx} \in \mathbb{R}^{K \times D}$.
Note that $\Sigma_{wx} = \Sigma_{xw}^T$ hence only one of these matrices has to be calculated from the samples in every training iteration.

Since these are high-dimensional matrices, obtaining a numerically stable sample estimate would require a large amount of data, which would require the use of a very large mini-batch. This constraint poses both a hardware problem (since standard GPUs cannot fit mini-batches of thousands of examples) and a potential generalization degradation (as smaller mini-batches have been linked to improved generalization \cite{Keskar_2017_Large_Batch_Training}).
To overcome this issue, similarly to batch normalization, we normalize both the logits derivative vector (which is our proxy for the weights), and the input images by their respective standard deviations along the batch dimension. By doing so, we improve the numerical stability of the estimated matrices towards the inversion in the following Lautum estimation.

We additionally use a standard exponentially decaying moving-average estimation of the three matrices $\Sigma_{w}, \Sigma_{wx}, \Sigma_{xw}$, to obtain numerical stability.
We denote by $\alpha$ the decay rate, and get the following update rule for the three covariance matrices in every training iteration:
\begin{align}
\label{eq:exp_sigma_sum}
    \Sigma_{(n)} = \alpha \Sigma_{(n-1)} + (1-\alpha) \Sigma_{batch},
\end{align}
where $n$ denotes the training iteration and $\Sigma_{batch}$ denotes the sample covariance matrix calculated using the current mini-batch. 
Using an exponentially decaying moving-average is of particularly high importance for $\Sigma_{w}$, which is inverted to compute the Lautum regularization term.

\noindent{\bf Training with Lautum regularization.}
Once the Lautum information has been estimated, our loss function for pre-transfer training is:
\begin{align}
\begin{split}
& Loss = \sum_{i=1}^{N} \sum_{k=1}^K -y_{ik}^{s} \log f_k(x_{i}^{s} | w_{\mathcal{D}}) - \lambda L(w_{\mathcal{D}}; x^{t}),
\end{split}
\end{align}
where the superscript $s$ represents the source dataset, and $t$ represents the target dataset.
Note that the cross-entropy loss is calculated using labeled samples from the source training set whereas the Lautum regularization uses only unlabeled samples from the target training set.
Also note that $y_i$ represents the ground truth label of the sample $x_i$; $f(x_{i} | w_{\mathcal{D}})$ represents the network's estimated post softmax label for that sample; and $L(w_{\mathcal{D}};x)$ is calculated as detailed above.
We emphasize that the Lautum regularization term is subtracted and not added to the cross-entropy loss as we aim at \emph{maximizing} the Lautum information during training.

The loss function for post-transfer training consists of a standard cross-entropy loss:
\begin{align}
\label{eq:post_transfer}
\begin{split}
& Loss = \sum_{i=1}^{N} \sum_{k=1}^K -y_{ik}^{t} \log f_k(x_{i}^{t} | w_{\mathcal{D}}).
\end{split}
\end{align}
Note that at this stage the cross-entropy loss, which is calculated using labeled target samples, implicitly includes the Lautum term of the target data (see Theorem~\ref{thm:test_ce_loss}).

\subsection{Mutual Information (MI) regularization in semi-supervised transfer learning}
\label{sec:Mutual_information_regularization}


We now turn to inspect the decomposition we obtained in Theorem \ref{thm:test_ce_loss_2}.
Firstly, the KL divergence term, $KL( p(y|x,w_{\mathcal{D}}) || f(y|x,w_{\mathcal{D}}) )$ and the conditional entropy term $H(x,y|w_{\mathcal{D}})$ can both be minimized where labels are available only.
Second, the term $H(y)$, which represents the intrinsic uncertainty of the labels, is inherent to the data and therefore cannot be minimized.
The last term, $I(x; w_{\mathcal{D}})$, does not depend on the task labels.
Therefore, by leveraging the unlabeled samples from the target dataset, it is possible to explicitly regularize the information that the model parameters hold with respect to the inputs, and thereby improve model transferability.

Another motivation for minimizing the information the model parameters hold with respect to the inputs stems from the information bottleneck (IB) theory \cite{tishby2000information}, which defines the IB Lagrangian by
\begin{equation}\label{IBLag}
\mathcal{L}(p(z|x)) = H(y|z) + \beta I(x;z).
\end{equation}

The IB Lagrangian addresses the relevance of the information in a representation via a tradeoff between its sufficiency to the task $H(y|z)$ )which is simply the cross-entropy loss), and its minimality $I(x;z)$, where the parameter $\beta$ allows tradeoff tuning.
By regularizing the learned representation $z$ to hold as minimal information as possible with respect to the inputs, we promote the representation to only hold information relevant to the task without overfitting the training data.

The regularization term in the IB is inherently different from our analysis in Theorem~\ref{thm:test_ce_loss_2}. The former relates the representation $z$ to the inputs, whereas our analysis relates the model weights to the inputs.
To bridge this difference we rely on the work in \cite{achille2018emergence}, which showed that $I(x;w_{\mathcal{D}})$ tightly bounds $I(x;z)+TC(z)$, which is the sum of the mutual information between the inputs and the representation and the total correlation of the representation $z$.
Combining this argument with Theorem~\ref{thm:test_ce_loss_2}, it follows that we may minimize the MI between the target inputs and their respective feature representations, thereby inducing the minimization of the MI between the target inputs and the network weights.
Relating this insight to the context of semi-supervised transfer learning, by using the unlabeled data of the target dataset, and regularizing the information the representations hold with respect to their inputs, we are able to optimize the cross-entropy loss of the target dataset.

\noindent {\bf Estimating the Mutual Information.}
Estimating the MI can be challenging, especially for high-dimensional distributions. To this end, we use the variational approach suggested in \cite{song2019understanding}, which is derived from the MINE method \cite{belghazi2018mine}.

The MINE \cite{belghazi2018mine} approach uses a neural network to estimate a lower bound for the Donsker-Varadhan representation \cite{donsker1983asymptotic} of the KL divergence. Plugging in the MI as defined in Definition~\ref{def:MI}, we get a lower bound for the MI value. The expressive power of a neural network enables this estimator to both be consistent and tightly estimate the MI.
The MINE bound reads as follows:
\begin{equation}
\label{KLlowB}
KL(P,Q) \geq \sup_{M \in \mathcal{F}}(\mathbb{E}_P[M] - log(\mathbb{E}_Q[e^M]),
\end{equation}
where $\mathcal{F}$ is a class of functions (neural networks in our case), $M : \Omega \xrightarrow{} \mathbb{R}$ is a specific neural network defined by the selected architecture and its parameters, $P$ is the empirical $p(x,z)$ which is the joint distribution of the target dataset inputs $x$ and their respective representations $z$, $Q$ is the empirical product of the marginals $p(x)$, $p(z)$ which are derived from the empirical samples of the target dataset, and lastly the neural network features are $z_i=f_{w_{\mathcal{D}}}(x_i)$.

Note that even though the MINE \cite{belghazi2018mine} estimator is unbiased, its variance is proportional to the MI value. Therefore, training the MI neural estimator in cases where the true MI value is large may lead to an unstable training and noisy results.
To mitigate this, we use the suggested estimator from \cite{song2019understanding}, which is biased and controls the bias-variance tradeoff using a clipping parameter.
Namely, the $\log(\mathbb{E}_Q[e^M])$ term from the lower bound in~\eqref{KLlowB} is modified to $log(\mathbb{E}_Q[Clip(e^M, e^{-\tau}, e^\tau)])$. Tuning the clipping value $\tau$ allows one to tradeoff the bias and variance.

\noindent {\bf Training with Mutual Information regularization.}
The estimated MI is used in the pre-transfer learning stage via a regularization term added to the cross-entropy loss, yielding the following optimization function for training:
\begin{align}
\begin{split}
& Loss = \sum_{i=1}^{N} \sum_{k=1}^K -y_{ik}^{s} \log f_k(x_{i}^{s} | w_{\mathcal{D}}) + \lambda I(x^{t};z^{t}).
\end{split}
\end{align}

where in this case as well, the superscript $s$ represents the source dataset, and $t$ represents the target dataset. The hyper-parameter $\lambda$ is optimized as part of our experiments.
Similarly to the Lautum regularization, we aim at minimizing the estimated value of the MI only for the part of the target dataset for which labels are not available.

Note that there is an interplay between the MI estimation and the learned classification function in the training process; the MI value stems from the MI neural estimator which makes use of the output features of the classification network. At the same time, training the classification network requires the MI estimation to calculate the regularization term. To address this training inter-dependency, both networks are trained conjointly. In order to facilitate a stable training process, we alternately train the two networks, i.e. we train the MI estimator for a few epochs, and then train the classification network for a few epochs. This process is repeated until convergence.

The post-transfer-learning stage follows the same process as in the Lautum case, i.e., the loss in \eqref{eq:post_transfer} is employed. In this case as well, MI regularization is not explicitly used but implicitly encapsulated in the cross-entropy loss.

\subsection{Comparing the Mutual Information and the Lautum Information regularizations}
\label{subsec:comparing_MI_and_Lautum}
Both MI regularization and Lautum Information regularization aim at controlling the information that the model parameters hold with respect to the inputs. However, each of them has an advantage over the other.

One fundamental difference is related to the assumptions that each method makes about the data: Lautum regularization assumes that the inputs and model weights are jointly Gaussian, whereas MI makes no such assumptions about either the data or model parameter distributions.

Another difference is that the Lautum regularization aims at maximizing an unbounded regularization term, a fact that may introduce a stability challenge in the training process and hyperparameter tuning. In contrast, MI regularization minimizes a bounded quantity, which makes the training process significantly more stable.

Computational complexity is a relevant differentiating factor as well. In cases of large inputs, Lautum regularization involves the computation of very large covariance matrices and their inversion, which requires heavy computing resources and therefore slows down training. 

Remarkably, one key disadvantage MI regularization has compared to Lautum regularization lies in the estimated regularization term itself: the Lautum Information approach estimates the value of the Lautum quantity itself, whereas the MI approach estimates a lower bound of the regularization quantity.
Therefore, MI regularization has an implicit impact on the value we actually aim at controlling, contrary to Lautum regularization's explicit impact.

Note that although the MI regularization holds several key advantages compared to the Lautum information regularization, including superior experimental results as outlined in Section~\ref{sec:Experiments}, we still find the latter a relevant case study. It paves a path for an information-theoretic-led analysis while also serving as an experimental stepping-stone in our research.

\section{Experiments}
\label{sec:Experiments}
In order to demonstrate the advantages of semi-supervised transfer learning both with the Lautum regularization and with the MI regularization, we performed several experiments on image classification tasks using deep neural networks.
Note, though, that our theoretical derivations also apply to other machine learning models.

\subsection{Experimental setup}
\label{subsec:Experimental setup}
We train deep neural networks and perform transfer learning from the original source dataset to the target dataset.
In our experiments we use the original labeled source training set as is, and split the target training set into two parts. The first part is very small and contains labeled samples, whereas the second part consists of the remainder of the target training set and contains unlabeled samples only (the labels are discarded).
The performance is evaluated by the final accuracy on the target test set.

We examined different methods of transfer learning:
(1) standard supervised transfer which uses the labeled samples only.
(2) Temporal Ensembling semi-supervised learning as outlined in \cite{Laine_18_TemporalEnsembling}, applied in a transfer learning setting; Temporal Ensembling is applied in the post-transfer training stage.
(3) Mk-MMD \cite{Mk_MMD}, which is based on 19 different Gaussian kernels with different standard deviations. Mk-MMD is applied in the pre-transfer training stage.
(4) Lautum regularization - our technique as described in Section~\ref{sec:Lautum information based semi-supervised transfer learning}.
(5) Both Temporal Ensembling and Lautum regularization. Note that Temporal Ensembling is applied in the post-transfer training stage, whereas Lautum regularization is applied in the pre-transfer training stage.
(6) MI regularization, our second technique which is described in Section \ref{sec:Mutual_information_regularization}.
(7) Both Lautum regularization and MI regularization applied in conjunction.

As presented in Sections~\ref{sec:Lautum information based semi-supervised transfer learning} and \ref{sec:Mutual_information_regularization} our training paradigm consists of two stages.
When employing our proposed methods, in the first stage we train the network using our fully labeled source training set while imposing either Lautum or MI or both regularizations using the unlabeled samples from the target training set.
We use the same mini-batch size both for the calculation of the cross-entropy loss (using labeled source samples) and for the computation of the regularization terms (using unlabeled target samples).

In the Lautum experiments, we also use an exponentially decaying moving average to ascertain numerical stability in the estimation of the covariance matrices $\Sigma_w, \Sigma_{wx}, \Sigma_{xw}$. Note that the matrix $\Sigma_x$ is calculated once before training and remains constant throughout it. 
In the MI experiments we train the MI neural estimator alternately with the classification network training. We empirically observed that this methodology provides a more stable training process, as particularly in our case we have a mutual dependency between the MI estimator training, which relies on the extracted features from the classifier, and the classifier training, which is regularized using the MI estimation. 

In the second stage, we perform a transfer to the target set by training (fine-tuning) the entire network using the labeled samples from the target training set, where the mini-batch size remains the same as before.
As in \cite{Zhou_2018_SSL_TL}, we fine-tune the entire network as this approach best fits the settings of semi-supervised transfer learning. As previously mentioned, we do not apply any information regularization at this stage as we empirically found it does not improve the obtained performance results.

We performed our experiments on the MNIST and CIFAR-10 datasets.
For MNIST we examined the transfer to the notMNIST dataset, which consists of 10 classes representing the letters A-J. The notMNIST dataset is similar to the MNIST dataset in its grayscale styling and image size, yet it differs in content.
For CIFAR-10 we examined the transfer to 10 specific classes of the CIFAR-100 dataset (specifically, classes 0, 10, 20,...,90 of CIFAR-100, which we reclassified as classes 0, 1, 2,...,9 respectively).
These CIFAR-100 classes are different from the corresponding CIFAR-10 ones in content. For example, class 0 in CIFAR-10 represents airplanes whereas class 0 in CIFAR-100 represents beavers.
Both in the MNIST $\rightarrow$ notMNIST case and the CIFAR-10 $\rightarrow$ CIFAR-100 (10 classes) case we used the same CNN as in \cite{Laine_18_TemporalEnsembling}. Appendix~\ref{sec:The CNN architecture used in the experiments} details the network architecture.

\subsection{MNIST to notMNIST results}
\label{subsec:MNIST to notMNIST results}
In order for the input images to fit the network's input we resized the MNIST and notMNIST images to $32$x$32$ pixels and transformed each of them into RGB format.
Training was done using the Adam optimizer \cite{Kingma15Adam} and a mini-batch size of 50 inputs. Standard supervised training with this network on the entire MNIST dataset lead to a test accuracy of $99.01\%$ on MNIST.

For each of the transfer learning methods we examined three different splits of the notMNIST training dataset which consists of 200,000 samples to an unlabeled part and a labeled part: (1) unlabeled part of 199,950 samples and a labeled part of 50 samples; (2) unlabeled part of 199,900 samples and a labeled part of 100 samples; (3) unlabeled part of 199,800 samples and a labeled part of 200 samples. All three options use very few labeled samples in order to fairly represent realistic semi-supervised learning scenarios; in all three options 99.9\% or more of the training data is unlabeled.


In the Lautum regularization case, we used a decay rate of $\alpha = 0.999$ for the exponentially decaying moving average estimation of the three matrices $\Sigma_w, \Sigma_{wx}, \Sigma_{xw}$.
Further comprehensive hyper-parameter tuning (e.g. for the value of $\lambda$ which controls the regularization term weight) was done for all examined methods.

Using the settings outlined above we obtained the results shown in Table~\ref{table:MkMMD_TE_Lautum_Comparison_MNIST} for the MNIST $\rightarrow$ notMNIST case.
The advantage of using information-based regularization is evident from the results, as it outperforms the other compared methods in all the examined target training set splits.
In general, the Temporal Ensembling method by itself does not yield very competitive results compared to standard transfer learning.
However, the MI regularization provides the best target test set accuracy over all the examined target training set splits.
Though our analysis in Section~\ref{sec:The Cross-Entropy Loss - an Information Theory Perspective} does not establish theoretical foundations to support using both MI and Lautum regularizations in conjunction, we performed this experiment to present a comprehensive empirical setup.
Our experiments show that imposing both regularizations degrades the target test accuracy compared to their individual applications.

In addition, we examined the contributions of Lautum and MI regularizations with respect to the number of labeled target training samples even further.
We tested several additional ratios of labeled target training samples up to 2.5\%, and compared our suggested methods to standard transfer learning.
The results are shown in Table~\ref{table:number_of_samples_impact}.
As the proportion of labeled samples in the target training set increases, the contribution of the regularizations is diminished as expected. This behavior is explained by Theorem~\ref{thm:test_ce_loss} and Theorem~\ref{thm:test_ce_loss_2}, as in both decompositions the role of the KL divergence term is enhanced by the increased number of labeled samples during cross-entropy loss minimization.

\begin{table}[ht!]
\begin{center}
\begin{tabular}{|c|c|c|}
 \hline
   \textbf{Method} &
   \textbf{\# labeled} &
   \textbf{Accuracy} \\ [0.5ex]
   \hline \hline
   Standard & 50 & 34.02\%  \\
   TE & 50 & 37.28\% \\
   Mk-MMD & 50 & 46.72\% \\
   Lautum + MI & 50 & 46.51\% \\
   Lautum & 50 & 48.42\% \\
   Lautum + TE & 50 & 67.3\% \\
   MI & 50 & 70.50\% \\
   \hline
   Standard & 100 & 57.58\% \\
   TE & 100 & 61.45\% \\
   Mk-MMD & 100 & 63.32\% \\
   Lautum + MI & 100 & 62.01\% \\
   Lautum & 100 & 65.94\% \\
   Lautum + TE & 100 & 76.8\% \\
   MI & 100 & 78.38\% \\
   \hline
   Standard & 200 & 67.78\% \\
   TE & 200 & 74.87\% \\
   Mk-MMD & 200 & 80.35\% \\
   Lautum + MI & 200 & 76.31\% \\
   Lautum & 200 & 82.8\% \\
   Lautum + TE & 200 & 84.5\% \\
   MI & 200 & 85.85\% \\
  \hline
\end{tabular}
\end{center}
\caption{MNIST $\rightarrow$ notMNIST, target test set accuracy comparison between various methods. Results are presented for different amounts of labeled target training samples.
}
\label{table:MkMMD_TE_Lautum_Comparison_MNIST}
\end{table}

\begin{table}[ht!]
\begin{center}
\begin{tabular}{|c|c|c|c|}
 \hline
   \textbf{\# labeled} &
   \textbf{Standard TL} &
   \textbf{Lautum} &
   \textbf{MI} \\ [0.5ex]
   \hline \hline
   50 & 34.02\% & 48.42\% & 70.50\% \\
   100 & 57.58\% & 65.94\% & 78.38\% \\
   200 & 67.78\% & 82.8\% & 85.85\% \\
   500 & 75.01\% & 87.91\% & 89.40\% \\
   1000 & 81.26\% & 90.04\% & 90.53\% \\
   2500 & 86.10\% & 91.85\% & 92.64\% \\
   5000 & 89.40\% & 93.78\% & 94.45\% \\
   \hline
\end{tabular}
\end{center}
\caption{MNIST $\rightarrow$ notMNIST, the contirbution of (i) Lautum regularization, and (ii) MI regularization, to the target test set accuracy as labeled sample size increases.}
\label{table:number_of_samples_impact}
\end{table}

\subsection{CIFAR-10 to CIFAR-100 (10 classes) results}
\label{subsec:CIFAR-10 to CIFAR-100 (10 classes) results}

In the CIFAR-10 $\rightarrow$ CIFAR-100 (10 classes) case training was done using the Adam optimizer \cite{Kingma15Adam} and a mini-batch size of 100 inputs. Standard supervised training with this network on the entire CIFAR-10 dataset lead to a test accuracy of $85.09\%$ on CIFAR-10.

Our target set consists of 10 classes from the CIFAR-100 dataset. Accordingly, our training target set consists of 5,000 samples and our test target set consists of 1,000 samples.
We examined the same seven transfer learning techniques as in the MNIST $\rightarrow$ notMNIST case, where for each we examined three different splits of the CIFAR-100 (10 classes) training set into an unlabeled part and a labeled part: (1) unlabeled part of 4,900 samples and a labeled part of 100 samples; (2) unlabeled part of 4,800 samples and a labeled part of 200 samples; (3) unlabeled part of 4,500 samples and a labeled part of 500 samples. All three options use a small number of labeled samples in order to fairly represent realistic semi-supervised learning scenarios; in all three options 90\% or more of the data is unlabeled.
We used a decay rate of $\alpha = 0.999$ for the exponentially decaying moving average estimation of the three matrices $\Sigma_w, \Sigma_{wx}, \Sigma_{xw}$.
In this case as well, further comprehensive hyper-parameter tuning was done for all examined methods.


Using the settings outlined above we obtained the results shown in Table~\ref{table:MkMMD_TE_Lautum_Comparison_CIFAR} for the CIFAR-10 $\rightarrow$ CIFAR-100 (10 classes) case.
It is evident from the results that in the CIFAR-10 $\rightarrow$ CIFAR-100 (10 classes) case as well, using either Lautum regularization or MI regularization improves the post-transfer performance on the target test set and outperforms Temporal Ensembling and Mk-MMD.
These results are consistent with the MNIST $\rightarrow$ NotMNIST results, as the MI regularization outperforms all other methods in this case too. Particularly in the case of 100 labeled target training samples, the obtained performance boost is significantly larger than the other methods.



\begin{table}[ht!]
\begin{center}
\begin{tabular}{|c|c|c|}
 \hline
   \textbf{Method} &
   \textbf{\# labeled} &
   \textbf{Accuracy} \\ [0.5ex]
   \hline \hline
   Standard & 100 & 39.90\% \\
   TE & 100 & 42.20\% \\
   Mk-MMD & 100 & 45.30\% \\
   Lautum + MI & 100 & 43.14\% \\
   Lautum & 100 & 46.42\% \\
   Lautum + TE & 100 & 46.80\% \\
   MI & 100 & 53.10\% \\
   \hline
   Standard & 200 & 52.80\%  \\
   TE & 200 & 54.60\% \\
   Mk-MMD & 200 & 59.30\% \\
   Lautum + MI & 200 & 55.44\% \\
   Lautum & 200 & 60.70\% \\
   Lautum + TE & 200 & 61.40\% \\
   MI & 200 & 63.20\% \\
   \hline
   Standard & 500 & 64.50\% \\
   TE & 500 & 66.50\% \\
   Mk-MMD & 500 & 68.00\% \\
   Lautum + MI & 500 & 66.93\% \\
   Lautum & 500 & 70.60\% \\
   Lautum + TE & 500 & 71.05\% \\
   MI & 500 & 72.82\% \\
  \hline
\end{tabular}
\end{center}
\caption{CIFAR-10 $\rightarrow$ CIFAR-100 (10 classes), target test set accuracy comparison between various methods. Experimental results are presented for different amounts of labeled training target samples.}
\label{table:MkMMD_TE_Lautum_Comparison_CIFAR}
\end{table}

\section{Conclusions}
\label{sec:Conclusions}

We have proposed a new semi-supervised transfer learning approach for machine learning algorithms that are trained using the cross-entropy loss. Our approach is backed by information-theoretic derivations and exemplifies how one may make good use of unlabeled target set samples along with just a few labeled samples to improve model performance on the target dataset.
Our approach relies on two techniques: (i) the maximization of the Lautum information between unlabeled samples from the target set and an algorithm's learned features by using the Lautum information as a regularization term; (ii) the minimization of the Mutual information between unlabeled samples from the target set and a model's learned parameters by using the Mutual information as a regularization term.
As shown, regularizing either of these terms minimizes the cross-entropy test loss on the target dataset and thereby improves performance as indicated by our experimental results.

The advantages of the MI regularization compared to the Lautum regularization are also supported by the experimental results where the performance of the former surpasses that of the latter in all tested settings.

We have also shown that our approaches surpass the performance of prominent semi-supervised techniques in a transfer learning setting.

Future work will focus on alternative approximations of the used information-theoretic quantities (Lautum information and MI) which could potentially yield better performance or reduce the introduced computational overhead.
In addition, our approach has the potential to be applied to other tasks as well, such as multi-task learning or domain adaptation.
Incorporating new techniques to mitigate the effects of training using an imbalanced dataset could also be of interest.
We defer these directions to future research.

\ifCLASSOPTIONcompsoc
  \section*{Acknowledgments}
\else
  \section*{Acknowledgment}
\fi
This work was supported by the ERC-StG SPADE grant.

\ifCLASSOPTIONcaptionsoff
  \newpage
\fi

\bibliographystyle{IEEEtran}
\bibliography{egbib}

\appendices

\section{Proof of Theorem 1}
\label{sec:Proof of Theorem 1}

Let us reiterate Theorem 1 before formally proving it.

\noindent \textbf{Theorem 1}
\emph{For a classification task with a ground-truth distribution $p(y|x)$, fixed training set $\mathcal{D}$, learned weights $w_\mathcal{D}$ and learned classification function $f(y|x, w_\mathcal{D})$, the expected cross-entropy loss of a machine learning algorithm on the test distribution is approximately}
\begin{equation}
    \label{eq:CE_test_loss_expression_appendix}
    \mathbb{E}_{w_{\mathcal{D}}} \left\{ KL( p(x,y) || f(x,y|w_{\mathcal{D}}) ) \right\} + H(y|x) - L(w_{\mathcal{D}};x).
\end{equation}

\noindent \textbf{Proof.}
The expected cross-entropy loss of the learned classification function $f(y|x, w_\mathcal{D})$ on the test distribution $p(x,y)$ can be approximated by
\begin{equation}
    \label{eq:Test_CE_appendix}
    \mathbb{E}_{(x,y) \sim p(x,y)} \mathbb{E}_{w \sim p(w_\mathcal{D})} \{-\log f(y|x,w_\mathcal{D})\},
\end{equation}
where instead of taking the expectation over $p(x,y,w_\mathcal{D})$, we take the expectation over $p(x,y)$ and $p(w_\mathcal{D})$ separately. Our approximation is motivated by the fact that the test distribution $p(x,y)$ is independent from $p(w_\mathcal{D})$ as post-training $w_\mathcal{D}$ is no longer a random variable but rather a fixed parameter of the problem.
Explicitly, \eqref{eq:Test_CE_appendix} can be written as
\footnote{Since the values of $y$ are discrete it is more accurate to sum instead of integrate over them. Yet, for the simplicity of the proof we present the derivations using integration.}
\begin{equation}
    \label{eq:full_expression_appendix}
    - \iiint p(x,y) p(w_\mathcal{D}) \log f(y|x,w_\mathcal{D}) \hspace{0.05 cm} dx \hspace{0.05 cm} dy \hspace{0.05 cm} dw_\mathcal{D}.
\end{equation}

To compare the learned classifier with the true classification of the data we develop \eqref{eq:full_expression_appendix} further as follows:
\begin{equation}
    = - \iiint p(x,y) p(w_\mathcal{D}) \log \left\{ \frac{f(y|x, w_\mathcal{D})}{p(y|x, w_\mathcal{D})} p(y|x, w_\mathcal{D}) \right\} \hspace{0.05 cm} dx \hspace{0.05 cm} dy \hspace{0.05 cm} dw_\mathcal{D}.
\end{equation}

Using standard logarithm arithmetic we get the following expression:
\begin{equation}
    \label{eq:Two_expressions}
    \begin{split}
     = & \underbrace{- \iiint p(x,y) p(w_\mathcal{D}) \log \left\{ \frac{f(y|x, w_\mathcal{D})}{p(y|x, w_\mathcal{D})} \right\} \hspace{0.05 cm} dx \hspace{0.05 cm} dy \hspace{0.05 cm} dw_\mathcal{D}}_{(\star)} \\
    & \underbrace{- \iiint p(x,y) p(w_\mathcal{D}) \log p(y|x, w_\mathcal{D}) \hspace{0.05 cm} dx \hspace{0.05 cm} dy \hspace{0.05 cm} dw_\mathcal{D}}_{(\star \star)}.
    \end{split}
\end{equation}

We separate the derivations of the two terms in \eqref{eq:Two_expressions}.
First, we develop the term $(\star \star)$ further:
\begin{equation}
    \begin{split}
    (\star \star) = &
    - \iiint p(x,y) p(w_{\mathcal{D}}) \log p(y|x,w_{\mathcal{D}}) \hspace{0.05 cm} dx \hspace{0.05 cm} dy \hspace{0.05 cm} dw_\mathcal{D} \\
    = & - \iiint p(x,y)p(w_{\mathcal{D}}) \log \left\{ \frac{p(x,y,w_{\mathcal{D}})}{p(x,w_{\mathcal{D}})} \hspace{0.05 cm} \right\} dx \hspace{0.05 cm} dy \hspace{0.05 cm} dw_\mathcal{D}.
    \end{split}
\end{equation}

Using logarithm arithmetic and adding and subtracting terms we get:
\begin{equation}
\begin{split}
\label{eq:star_4_terms_appendix}
(\star \star) = 
& - \iiint p(x,y) p(w_{\mathcal{D}}) \log \left\{ \frac{p(x, y, w_{\mathcal{D}})}
{p(x,y) p(w_{\mathcal{D}})} \right\} dx \hspace{0.05 cm} dy \hspace{0.05 cm} dw_\mathcal{D}
\\
& - \iiint p(x,y) p(w_{\mathcal{D}}) \log \left\{ p(x,y) p(w_{\mathcal{D}})
\right\} dx \hspace{0.05 cm} dy \hspace{0.05 cm} dw_\mathcal{D}
\\
& - \iiint p(x,y) p(w_{\mathcal{D}}) \log \left\{ \frac{p(x) p(w_{\mathcal{D}})}
{p(x, w_{\mathcal{D}})} \right\} dx \hspace{0.05 cm} dy \hspace{0.05 cm} dw_\mathcal{D}
\\
& + \iiint p(x,y) p(w_{\mathcal{D}}) \log \left\{ p(x) p(w_{\mathcal{D}}) \right\} dx \hspace{0.05 cm} dy \hspace{0.05 cm} dw_\mathcal{D}.
\end{split}
\end{equation}

Using the law of total probability along with the definitions of the differential entropy and the Lautum information we can reformulate \eqref{eq:star_4_terms_appendix} as follows:
\begin{equation}
\begin{split}
    (\star \star)
    & = L(w_{\mathcal{D}}; (x,y)) \\
    & + H(w_{\mathcal{D}}) + H(x,y) \\
    & - L(w_{\mathcal{D}};x) \\
    & - H(x) - H(w_{\mathcal{D}}).
    \end{split}
\end{equation}

Since $H(y|x) = H(x,y) - H(x)$ we get that:
\begin{equation}
    \label{eq:star_final_exp}
    (\star \star) = L(w_{\mathcal{D}};(x,y)) + H(y|x) - L(w_{\mathcal{D}};x).
\end{equation}

We next analyze the expression of $(\star)$:
\begin{equation}
    \begin{split}
    (\star) = &
    - \iiint p(x,y) p(w_{\mathcal{D}}) \log \left\{ \frac{f(y|x,w_{\mathcal{D}})}{p(y|x,w_{\mathcal{D}})} \right\} \hspace{0.05 cm} dx \hspace{0.05 cm} dy \hspace{0.05 cm} dw_\mathcal{D}.
    \end{split}
\end{equation}

Within the $\log$ operation we multiply and divide by the term $\frac{p(x,y)p(w_{\mathcal{D}})}{p(x,w_{\mathcal{D}})}$ and get:
\begin{equation}
    \begin{split}
    & (\star) = \iiint p(x,y) p(w_{\mathcal{D}}) \cdot \\
    & \log \left\{
    \frac{p(x,y)p(w_{\mathcal{D}})}{f(y | x, w_{\mathcal{D}}) p(x,w_{\mathcal{D}})}
    \cdot 
    \frac{p(y | x, w_{\mathcal{D}}) p(x,w_{\mathcal{D}})}{p(x,y)p(w_{\mathcal{D}})}
    \right\}
    dx \hspace{0.05 cm} dy \hspace{0.05 cm} dw_\mathcal{D}.
    \end{split}
\end{equation}

Since $f(y|x,w_{\mathcal{D}})$ is the learned classifier which outputs the probability of the label $y$ for an input $x$ given the model weights $w_{\mathcal{D}}$, without the labels it has no affect on the joint distribution of the weights and inputs, i.e. $f(x,w_{\mathcal{D}}) = p(x,w_{\mathcal{D}})$, $f(w_{\mathcal{D}}) = p(w_{\mathcal{D}})$. Accordingly,
\begin{equation}
\begin{split}
    (\star) = & \iiint p(x,y) p(w_{\mathcal{D}}) \log \left\{ \frac{p(x,y)p(w_{\mathcal{D}})}{f(x,y,w_{\mathcal{D}})} \right\} dx \hspace{0.05 cm} dy \hspace{0.05 cm} dw_\mathcal{D} \\
    - & \iiint p(x,y) p(w_{\mathcal{D}}) \log \left\{ \frac{p(x,y)p(w_{\mathcal{D}})}{p(x,y,w_{\mathcal{D}})} \right\} dx \hspace{0.05 cm} dy \hspace{0.05 cm} dw_\mathcal{D}.
\end{split}
\end{equation}

Since $f(x,y|w_{\mathcal{D}}) = \frac{f(x,y,w_{\mathcal{D}})}{p(w_{\mathcal{D}})}$ we get that:
\begin{equation}
\begin{split}
    (\star) = & \iiint p(x,y) p(w_{\mathcal{D}}) \log \left\{ \frac{p(x,y)}{f(x,y|w_{\mathcal{D}})} \right\} dx \hspace{0.05 cm} dy \hspace{0.05 cm} dw_\mathcal{D} \\
    - & \iiint p(x,y) p(w_{\mathcal{D}}) \log \left\{ \frac{p(x,y)p(w_{\mathcal{D}})}{p(x,y,w_{\mathcal{D}})} \right\} dx \hspace{0.05 cm} dy \hspace{0.05 cm} dw_\mathcal{D}.
\end{split}
\end{equation}
In the first term we have the expectation over $w_\mathcal{D}$ and so:
\begin{equation}
\begin{split}
    (\star) = &  \mathbb{E}_{w_{\mathcal{D}}} \left\{ \iint p(x,y) \log \left\{ \frac{p(x,y)}{f(x,y|w_{\mathcal{D}})} \right\} dx \hspace{0.05 cm} dy \right\} \\
    - & \iiint p(x,y) p(w_{\mathcal{D}}) \log \left\{ \frac{p(x,y)p(w_{\mathcal{D}})}{p(x,y,w_{\mathcal{D}})} \right\} dx \hspace{0.05 cm} dy \hspace{0.05 cm} dw_\mathcal{D}.
\end{split}
\end{equation}

We get that the first term is the expectation over $w_\mathcal{D}$ of the KL-divergence between $p(x,y)$ and $f(x,y|w_{\mathcal{D}})$, whereas the second term is the negative Lautum information between $(x,y)$ and $w_{\mathcal{D}}$:
\begin{equation}
\begin{split}
    \label{eq:star_star_final_exp}
    (\star) = \mathbb{E}_{w_{\mathcal{D}}} \left\{ KL( p(x,y) || f(x,y|w_{\mathcal{D}}) ) \right\}
    - L(w_{\mathcal{D}};(x,y)).
\end{split}
\end{equation}

Plugging the expressions we got for $(\star \star)$ from \eqref{eq:star_final_exp} and for $(\star)$ from \eqref{eq:star_star_final_exp} into \eqref{eq:Two_expressions} we obtain the expression in \eqref{eq:CE_test_loss_expression_appendix}:
\begin{equation}
    \label{eq:final_term}
    \begin{split}
    & (\star) \hspace{0.05cm} + \hspace{0.05cm} (\star \star) =
    \underbrace{\mathbb{E}_{w_{\mathcal{D}}} \left\{ KL( p(x,y) || f(x,y|w_{\mathcal{D}}) ) \right\} - L(w_{\mathcal{D}};(x,y))}_{(\star)} \\
    & + \underbrace{L(w_{\mathcal{D}};(x,y)) + H(y|x) - L(w_{\mathcal{D}};x)}_{(\star \star)} \\
    & = \mathbb{E}_{w_{\mathcal{D}}} \left\{ KL( p(x,y) || f(x,y|w_{\mathcal{D}}) ) \right\} + H(y|x) - L(w_{\mathcal{D}};x).
    \end{split}
\end{equation}

\section{Proof of Theorem 2}
\label{sec:Proof of Theorem 2}
Let us reiterate Theorem 2 before formally proving it.

\noindent \textbf{Theorem 2}
\emph{For a classification task with a ground-truth distribution $p(y|x)$, fixed training set $\mathcal{D}$, learned weights $w_\mathcal{D}$ and learned classification function $f(y|x, w_\mathcal{D})$, the expected cross-entropy loss of a machine learning algorithm on the test distribution is equal to}
\begin{align}
\begin{split}
    & \mathbb{E}_{x, w_{\mathcal{D}}} \left\{ KL( p(y|x,w_{\mathcal{D}}) || f(y|x,w_{\mathcal{D}}) ) \right\} + \\
    & H(x,y|w_{\mathcal{D}}) - H(y) + I(x;w_{\mathcal{D}}).
\end{split}
\end{align}

\noindent \textbf{Proof.}
Recall that the cross-entropy can be written as:
\begin{equation}
    H(p,q)=H(p)+KL(p,q)
\end{equation}
In our notation:
\begin{equation}
    H(y\lvert x,w_{\mathcal{D}})+\mathbb{E}_{x,w_{\mathcal{D}}}\left\{KL(p(y\lvert x,w_{\mathcal{D}})||f(y\lvert x,w_{\mathcal{D}})) \right\}.
\end{equation}
Using the entropy chain rule we can write the left term as:
\begin{equation}
    H(y\lvert x,w_{\mathcal{D}})=H(y,x,w_{\mathcal{D}})-H(y)-H(w_{\mathcal{D}} \lvert x).
\end{equation}
From the following identity of MI, we can rewrite the relative entropy term: 
\begin{equation}
    \begin{aligned}
        I(x;w_{\mathcal{D}})&=H(w_{\mathcal{D}})-H(w_{\mathcal{D}} \lvert x) \Rightarrow\\
        H(w_{\mathcal{D}} \lvert x)&=H(w_{\mathcal{D}})-I(x;w_{\mathcal{D}})
    \end{aligned}
\end{equation}
In addition, the joint Entropy satisfies
\begin{equation}
    H(y,x,w_{\mathcal{D}})=H(w_{\mathcal{D}})+H(y,x \lvert w_{\mathcal{D}}).
\end{equation}
Therefore, we have that
\begin{equation}
    H(y\lvert x,w_{\mathcal{D}})=H(w_{\mathcal{D}})+H(y,x \lvert w_{\mathcal{D}})-H(y)+I(x;w_{\mathcal{D}})-H(w_{\mathcal{D}}),
\end{equation}
which leads to our final cross-entropy decomposition:
\begin{align}
\begin{split}
    & \mathbb{E}_{x, w_{\mathcal{D}}} \left\{ KL( p(y|x,w_{\mathcal{D}}) || f(y|x,w_{\mathcal{D}}) ) \right\} + \\
    & H(x,y|w_{\mathcal{D}}) - H(y) + I(x;w_{\mathcal{D}}).
\end{split}
\end{align}

\begin{figure}[t]
    \centering
    \includegraphics[width=\columnwidth]{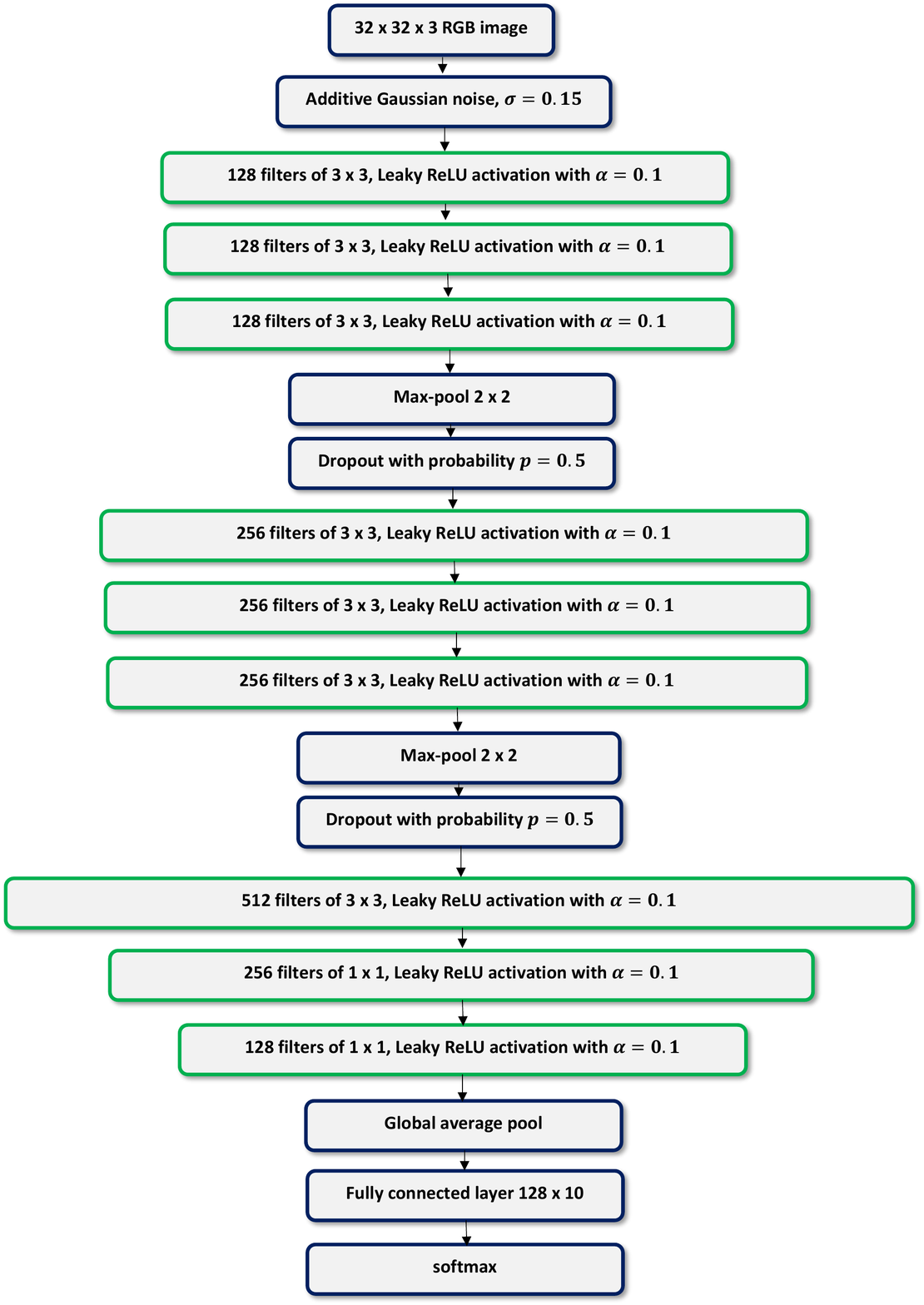}
    \caption{The network architecture used in our experiments.} \label{fig:Network_architecture}
\end{figure}

\section{The CNN architecture used in the experiments}
\label{sec:The CNN architecture used in the experiments}

Both in the MNIST $\rightarrow$ notMNIST case and the CIFAR-10 $\rightarrow$ CIFAR-100 (10 classes) case we used the same CNN as in \cite{Laine_18_TemporalEnsembling}. The architecture is illustrated in Figure~\ref{fig:Network_architecture}.

\end{document}